# Supporting Navigation of Outdoor Shopping Complexes for Visually-impaired Users through Multi-modal Data Fusion

Archana Paladugu, Parag S. Chandakkar, Peng Zhang and Baoxin Li

Computer Science and Engineering
Arizona State University
{apaladug,pchandak,pzhang41,baoxin.li}@asu.edu

## ABSTRACT

Outdoor shopping complexes (*OSC*) are extremely difficult for people with visual impairment to navigate. Existing GPS devices are mostly designed for roadside navigation and seldom transition well into an OSC-like setting. We report our study on the challenges faced by a blind person in navigating OSC through developing a new mobile application named *iExplore*. We first report an exploratory study aiming at deriving specific design principles for building this system by learning the unique challenges of the problem. Then we present a methodology that can be used to derive the necessary information for the development of iExplore, followed by experimental validation of the technology by a group of visually impaired users in a local outdoor shopping center. User feedback and other experiments suggest that iExplore, while at its very initial phase, has the potential of filling a practical gap in existing assistive technologies for the visually impaired.

## 1. INTRODUCTION

Navigating an unfamiliar outdoor space is a challenging task for visually impaired (*VI*) people. The use of technological devices in conjunction with specialized training and accessible architecture of buildings and roads has made this task possible. However, there are still many factors that make this task stressful and sometimes dangerous for VI users. The situation is worsened especially for outdoor shopping complexes (*OSC*), whose building layout is often cluttered with haphazard placement of stores, parking spaces, walkways and driving roads, etc (Fig. 1). Existing GPS devices seldom provide useful navigation formation for a VI user in such environment .

   The first problem towards building a system for OSC navigation is the lack of awareness among developers regarding the inadequacy of the current systems in place and lack of studies stating ways to overcome the existing information deficit. This paper reports our study on the challenges faced by VI users when navigating OSC and lists the necessary information required to bridge this gap. Secondly, the required information is currently only available from one source. We propose a solution to fuse information from different sources. Thirdly, once the necessary information is in place, our application catering to VI users has presented a voice over for visual data. We evaluate these techniques and propose a system design that takes after the design paradigms obtained from our user studies. This paper delves into these three aspects that are required for the development of this system.

   The focus on OSC was motivated both by their direct relevance to the life of VI users and by the fact that conventional map services do not provide navigation for an OSC. Typical OSC are fairly complex in layout and thus serve well as representatives of general outdoor building complexes. Besides building a functional iPhone app that is freely available to our participants, the work contributes to developing general principles and guidelines in designing interface schemes and information representation paradigms for supporting mobile-device-based navigation assistance for people with visual impairment. The paper is organized as follows. The literature is presented in Section 2. We present the user survey on the current deficit and present our assessment of required information in Section 3. We present our proposed approach covering details of data fusion, design of the interface and schema used for data presentation in Section 4. Our evaluation is presented in Section 5. The conclusion and discussion is presented in Section 6.

## 2. RELATED WORK

Navigation assistance for visually impaired users includes accessible infrastructure, specialized orientation and mobility training [1], and technological aids [2]. Technological aids, which are the focus of this work, may be devices that give live help on site or tools that help the user to prepare for the journey ahead of time. A review of dedicated devices for on-site navigation help using GPS, sonar, laser, RFID, etc. can be found in [3]. Recent years have seen new systems that are built upon general-purpose touchscreen mobile devices, moving away from the requirement of dedicated hardware. The most dominant players of this type are these three iPhone apps: Ariadne GPS (www.ariadnegps.eu), MotionX GPS (gps.motionx.com) and Lookaround GPS (www.senderogroup.com). Google's Intersection Explorer allows touch-based walk able path exploration. Recent years have seen add-on techniques to make the underlying maps more helpful by using audio, haptic [8] and spatial tactile feedback [7]. There are relatively fewer methods/systems for supporting exploration and spatial learning of locations (to help a user's planning of a trip ahead of time). Examples include tactile map systems [4] and verbal description generation methods [5].

Despite the existence of the aforementioned work, the reality remains that none of them can support navigation in an OSC setting largely due to the lack of adequate mapping information [6], even with the newer type of systems like Ariadne GPS. A solution to this deficit is to obtain information for various sources and fuse them before integrating them into the application. Some prior work on map registration and geo-spatial data conflation addresses the problem of combining data from different sources to obtain the information for the applications that require additional geo spatial information. A method was proposed in [9] to align successive images taken aerially with an overall map of the region using feature based registration and mosaicking techniques. Linear features like active contours to register images were proposed in [13]. A technique that uses feature matching using RMS estimation on affine transform was proposed in [14]. Street data were used as control points and then cross correlation was employed to obtain matching in [15]. None of these methods are directly applicable for the following reasons. Firstly, these methods have prior knowledge of the structure of their maps and available information in their maps. Our data fusion uses information obtained from standard maps as well as shopping directories crawled from the web which can be viewed as pseudo-maps. Secondly, detecting control points now becomes an altogether different task due to the dissimilarity between two images on a lower-level. The data fusion for our application needs to be flexible enough to work with a reasonable accuracy for images obtained online with large variation in their structure. Further, current design of some of those systems is essentially based on the concept of making the underlying information (targeted at sighted users) more accessible to VI users, without taking into the real needs of the VI user in navigating a place.

The importance of having a system in place can be emphasized by the fact that Google and Bing maps are working on integrating floor plans of malls into their map service. Google Maps has a user interface [17] published for the user to identify three control points and scale the map on the top of their street view map for user input. Bing maps has also handled similar problem [18] without the need for human intervention on popular shopping malls. Still these map services are unable to provide navigation inside OSC, and there is a need for a framework with data fusion to truly build an application for a VI user in navigating OSC, which is the focus of our study.

## 3. USER SURVEY ON CHALLENGES FACED

We started an exploratory study with a group of visually impaired smart phone users in order to understand the challenges and identify deficits of existing solutions. . Feedback from our participants and online surveys of visually-impaired communities helped us to conclude that that the iPhone appears to be the most used and preferred smart phone among users with visual impairment. Most of the users are familiar with the voice-over feature on the phone. Three of the five users we interviewed reported that they collect information from the store before planning their visit. The information asked for includes directions from transit stop to the store entrance, landmarks that can identify the store, etc. The users reported a significant ease in navigation using GPS devices when walking on mainstream streets as opposed to walking inside an OSC. All our participants reported that seeking help from nearby humans would be their final resort. A summary of the key findings in the case study is given below:

- An on-demand description of their surroundings is always helpful for them to orient themselves.
- Description of the surroundings can be effectively done in terms of egocentric or allocentric methods. The usability and preference of this description varies widely from user to user and by location.
- Some users prefer to have the description of distances given in steps, as opposed to feet; low-vision users may still enjoy the availability of a zoomed map.
- Tactile landmarks are preferred to assert the location.
- Using angles for direction is not usable. But the users are familiar with terms like "diagonally right/left".
- It is not a good idea to direct them to walk through parking lots. It is often unsafe to do so and would involve hitting the cars with the cane.
- Longer safer routes are preferred over shorter routes with parking lots or obstacles.
- Extra information: Restroom location, traffic conditions on streets encountered, user created markers for future reference, etc. would be great add-ons.

Part of the study involved asking our participants to navigate a chosen local shopping complex with the help of the two aforementioned iPhone apps. It is to be noted that these apps helped only for navigating the major streets bordering the shopping complex but failed to provide much assistance for navigating the complex itself.

## 3.1. Design Guidelines for Building Apps for OSC

Based on our case study, we established a set of guidelines for developing an effective iPhone app in addressing the deficits of existing solutions. We summarize them below:

*Avoid additional screens as much as possible:* A VI user would face frustration in using finger gestures to get back and forth in a multi-layer menu and thus a flat structure should be used as much as possible in the interface.
*Less is more:* Our users disliked navigating through a page filled with too many buttons. They asked for a few functionalities that convey a lot of information, during our case study questionnaire.
*Layered information delivery:* Users with varying abilities will need different amount of information. Having information interlaced with gestures is preferred, so additional information is presented only on-demand.
*Orientation and Mobility training:* Most of the blind users have undergone the O&M training. The app needs to be consistent with the protocols in presenting the information.
*Supporting user notes:* Every user from our case study picked up some different cues about his/her surroundings. A mechanism for the user to record his/her own markings would enhance the usefulness of the application.
*Supporting user customization:* Low-vision and completely-blind users have different requirements. Users may prefer measuring distance in different ways. The app should allow some user-level customization to support such features.

## 4. PROPOSED DESIGN

We now present the design of a novel iPhone app that aims at addressing the challenges faced by VI users in navigating OSC. Largely based upon the design principles which were derived from the exploratory study, the proposed design and implementation attempts to overcome the challenges from the following four aspects: an information fusion technique, an information representation structure that is appropriate for mapping-related tasks in an OSC setting, an intuitive interface and interaction scheme, and the support for user-customization to cater to individual needs of the users. These are shown in the figure 1.

### 4.1. Tiered Information Creation

The major part of the problems faced when using map-based technologies for navigation in OSC is the lack of required information. Publicly available map services such as Google maps do not have the desired level of details for typical OSC. However, most shopping centers maintain and publish maps with rich annotations. Also, a sighted volunteer may be able to label a satellite image of a shopping complex as to where are the parking lots etc. Finally, a user while using the app, may want to insert his/her notes to a location. Considering all these possibilities, we adopt a tiered representation for all the mapping information. Figure 1 illustrates how this is currently implemented. In Figure 1, the base layer corresponds to the map that is typically available from a GPS system, the second layer is the layout map given by the shopping center, the third layer illustrates an image of the same locale with additional labels, and the fourth layer is used to store user notes.

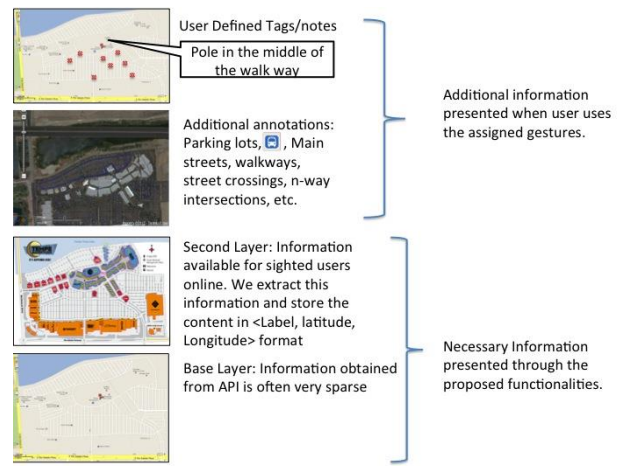

**Fig. 1.** The tiered information representation scheme used to support the application.

4.1.1. Base Layer Information

As seen in Figure 1, the base layer contains information available in the typical mobile applications using Google/Bing/Apple/other maps. This information is sometimes sparse, and from our studies, often inaccurate in an OSC setting. The locations of individual stores and bus stations are more often than not, inaccurate and haphazardly placed. Using this information for VI users is often dangerous for this reason. In our scheme, we find additional sources to obtain this kind of information with reasonable accuracy and map it on top of the base layer.

4.1.2. Second Layer: Information from the Web

The most vital information when navigating an OSC is the location of the stores inside the mall. This is the kind of information that is not available on the Google maps. One way to obtain this sort of information is to devise a method to integrate the Google maps with the store directories available on the website of OSC. . This can be viewed as a map-to-store map registration problem. We try to register the shopping mall directory to its corresponding Google map. The purpose is two-fold. Shopping directory has much more information as compared to maps and by overlaying shopping directory onto Google maps, we can still retain the GPS coordinate information.

Control point detection is a major step in any registration problem. We propose the use of shop centers as the control points as the shapes of stores in both Google maps and shopping directory have high-level similarity. We also employ road and parking lot detection but are not used for extracting control points since shopping directory may not have them. Firstly, we detect yellow-colored major roads and orange-colored freeways from Google maps by simple color segmentation. Black-colored text labels on Google maps are detected by using the same principle. Roads in Google Maps always have text labels on them and they have a lower-limit on their width. We use this fact to distinguish parking lots from roads. The roads are detected by using region-growing image segmentation technique in combination with distance transform. Distance transform allows us to monitor the width of the road. If it falls below the predefined minimum width or if there is no "white" road, then the region-growing stops. Text labels act as seed points while performing region-growing.

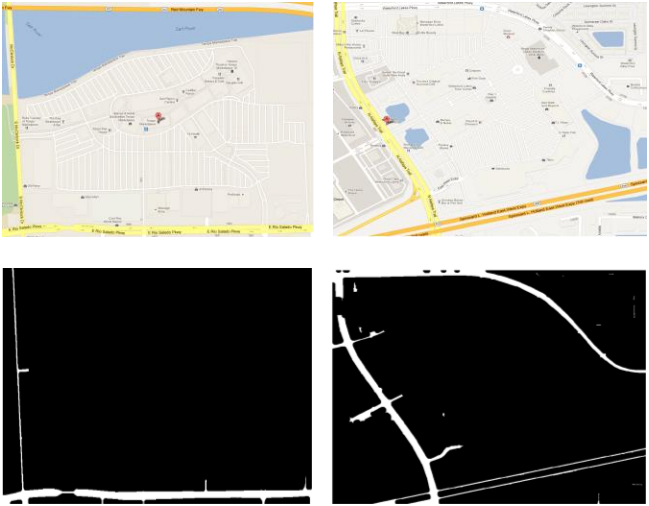

**Fig 2.** Robust street detection

The next procedure involves detecting stores in Google maps and the corresponding shopping directory. In Google maps, each shop has a label associated with it and it also has a border. Thus we can detect stores by extracting the entire portion which has the same color as the background color of the text label. The convex hull of the portion is calculated and those having arbitrary shape (not resembling to rectangle) are discarded. This is done based on the assumption that each shop has a regular shape. To detect stores from shopping directory, we ask user to select points which represent colors of the stores in the shopping directory. After this, the stores are detected by simple color segmentation. The control points in the two binary images are the centroids of each disconnected component (store). The centers are overlaid onto the original image for visualization purpose in figure 4.

Our problem inherently has lot of outliers because of "extra" information present in the shopping directory and thus we need an algorithm, which can perform better in presence of noise and outliers. Therefore we propose the use of Coheren

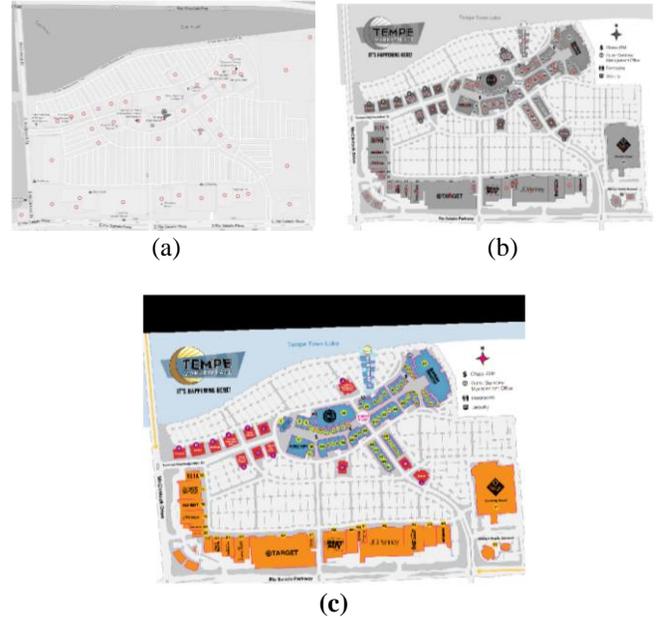

**Fig 3.** Control point detection in (a) Google maps and (b) shopping mall directory (c) Registration of Google map and shopping directory

Coherent Point Drift (CPD) algorithm proposed in [16]. It considers the registration problem as probability density estimation problem. It represents the first data point set by Gaussian Mixture Model (GMM) centroids and tries to fit the second data point set by maximizing the likelihood. The results obtained are indeed promising and they are shown below:

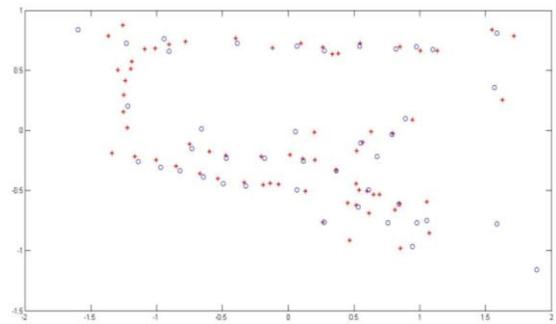

**Fig 4.** Registration of two sets of control points using CPD

4.1.3. Additional Annotations

According to our user consensus, they would like to know the location and the extent of parking lots, walkways and bus route information. This kind of metadata is not available on any mapping service. But the imagery of the maps shows a clear distinction that can be utilized to demarcate these areas. In our scheme, we do parking lot and walkway detection. The parking lot detection is carried out as described in the previous section. The walkways in front of stores can be detected by using the grey color gradient used consistently at this scale. We add bus route information obtained from the metadata to obtain all the additional data requested in the user survey.

4.1.4. User Defined Tags

In spite of providing additional user friendly information, there were some elements of each location that the visually impaired users highlighted as useful in navigation. This information ranged from the presence of water fountains, notes on location of shrubs on the walkways, narrowness of some passages, etc. Such information could only be gathered from the users themselves. We collected this information from our participants, orientation and mobility instructors and spouses of VI participants and geo tagged this information as a layer.

**4.2. Supporting User Customization**

The necessity of supporting user customization has been concluded earlier. In our current implementation, we support the following three types of customization. The first customization provided is the user preference on distance metric. Some users preferred the usage of steps and blocks to usage of absolute distance in terms of feet. We introduce a user-based calibration feature that calculates the step-to-feet ratio. The user is asked to walk 20 steps prior to the visit in a familiar environment to allow for calibration. The second customization is the method used to convey direction. Users prefer egocentric or allocentric method of description, depending on the location and complexity of the surroundings. The third customization provided is based on the level of vision of the user. The totally-blind users interact with the system using the predestined gestures and through voice-over output. The low-vision users can have the additional freedom of interacting with a spatial zoom-able map and larger font sizes.

**4.3. An Intuitive Interface for Supporting Necessary Functionalities**

Based on our case study, user input and our understanding of the existing GPS devices, we propose the following interface to help users navigating OSC. Our application uses iPhones accessibility mode and uses the standard voice-over gestures. A user can scroll though the buttons without knowing their spatial locations, using one finger swipe. A double tap anywhere on the screen selects an item. We refrain from introducing too many additional gestures to make the app as simple to use as possible.

*Homepage:* The homepage of the app consists of an entry into all the possible functions of the app. We adhered to the design paradigm inferred from the user study, stating the user preference of not having too many screens to navigate and having a few necessary functions on the screen. We have designed our homepage in such a way that the user can obtain all the important information by staying on the homepage. We interlaced various gestures to help access additional information. Figure 2 shows a screenshot of our homepage and an overview of all the functions.

*Where am I?* "Where am I" is a popular functionality provided by most of the existing GPS applications. A sample result for this function can be illustrated as follows: "Facing North near 661 Meadow Avenue". Once inside a shopping space, a description in terms of an address is no longer relevant information. We modify this functionality to make it more informative and convey necessary information in a tiered manner. When a user double taps on the "Where am I?" button, our application reads out the orientation of the user, the closest landmark and the nearest landmarks in either directions as shown in Figure 2. He repeats this gesture to listen to this information again. The user has an option to ask for more information after he listens to this information. A pre-assigned gesture (3 taps on the iPhone) provides information about the nearest streets and any user tagged notes if they exist around this location. He repeats this gesture to listen to this information again.

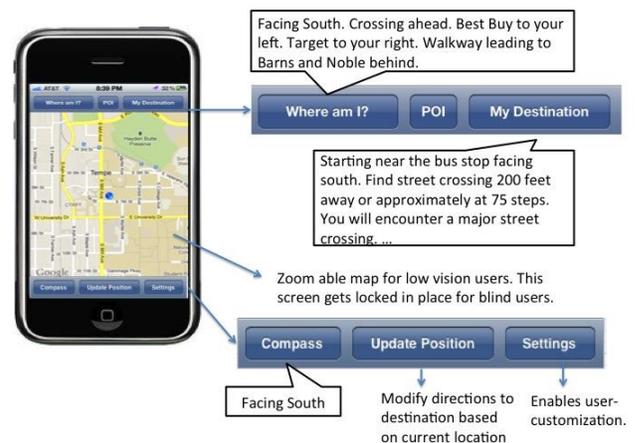

**Fig. 4.** An overview of iExplore's homepage and a sample output for each of the proposed functions can be seen above.

*Points of Interest:* Most applications and GPS devices support points of interest, where a predestined number of landmarks around the current location are listed out to the user in terms of distance and direction. We include this feature in our application, except that the information associated with this button is according to the tiered representation discussed above.

*Where is my destination?:* The directions provided by most applications are hard to use for VI users. We propose a scheme to provide blind-friendly directions. Once inside an OSC, the user can be standing inside a parking lot, a store or at a major landmark. Our description of the destination location takes into account that users do not like walking through the parking lots, considers the safety quotient of streets, includes description of places in-between in terms of stores lining the route and major streets on the way. The user inputs the name of the store or landmark he is interested in using the speech input feature provided. A verbal description (as seen in Figure 2) is then generated tailored to the user preference of egocentric or allocentric methodology, in terms of their distance preference. We use Dijkstra's algorithm to compute the shortest path among the walkable options. Using this path, our description takes into account relative positions of the landmarks and additional tagged information about the surroundings

## 5. EVALUATION

The proposed design has been implemented as an iPhone app called iExplore, which was tested by four participants for navigating a local outdoor shopping complex. We summarize some major outcomes of the tests below. Although this method is applicable to any outdoor shopping complex, for the sake of user studies, we picked one local mall and conducted our experiments at this location. The statistics of this experiment are presented below. The dataset consists of 6 outdoor shopping complexes. We used the additional features of the API to obtain street map with and without labels. We tested the accuracy of map registration and parking lot detection on this dataset. The dataset was chosen on the basis of availability of store map directories on their websites.

### 5.1. Accuracy of the Map Registration Technique

After registering Google maps to the corresponding shopping directory, we develop an evaluation scheme to validate proposed method. It is described as follows: We detect stores in Google maps as well as in the registered image. Now, we estimate the percentage of pixels of the Google map which overlap with the pixels of registered image. This also validates our shop detection algorithm. Figure 5(a) and 5(b) show the result of shop detection onto the Google map and the shopping directory. There is a 66.45% overlap in the area marked by our approach when compared to the original store directory and the ground truth. The parking lot detection is 100 percent accurate.

### 5.2. User Feedback on Overall System Design

Upon loading the app on the user's iPhone, we asked them to take time and get familiar with the buttons and tabs on the home screen. The users were asked to think aloud (make explicit comments) while using the app. They were not provided any additional training on the usage. After the users were familiar with all the functionalities, they were asked questions about the interface, its usability, the intuitiveness of each button, easiness of navigating the buttons, etc. The users were encouraged to ask us questions to clarify any aspect they found confusing. One key observation every participant made was that the interface has very few selectable items on the screen. They were able to summarize the key components upon closing the app. They also reported to have received more useful information than they would have hoped for (compared with other systems) from each functionally buttons.

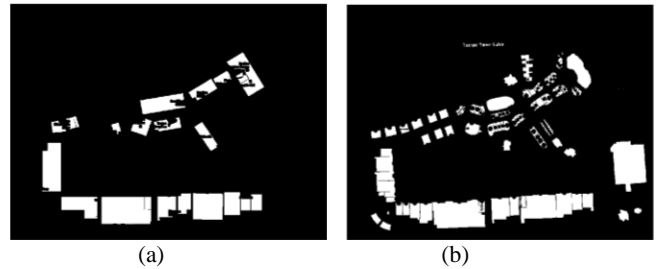

**Fig 5.** Store detection in (a) Google map (b) Registered image

### 5.3. Testing the Proposed Functionalities

The users were taken back to the testing site, where we conducted our initial case studies. We started at the bus station for uniformity and to simulate a real-time scenario. The users were asked to test each of the functionalities and were asked to walk around. We tagged them with a volunteer for safety reasons and the volunteer collected feedback from the users. The user was asked to change the settings to his preference of distance, terminology for directions, etc. Once done, the user is asked to go back to the home screen. He is asked to find the "Where am I" button and access the information. We then asked the user to point out the direction using their hand, towards one of the landmarks mentioned. This test was aimed at testing out the accuracy of information conveyed as well as user understanding from the verbal descriptions provided. The users got their directions right with 100% accuracy.

### 5.4. Evaluating Navigation Support

To evaluate the support provided by iExplore for navigation, we defined preset routes with 2 turns and asked the users to use the app to find the destination. If the participant took a wrong turn, we recorded a miss for that segment of the journey and let the user turn around. The user was asked to use the app to gain a better understanding of his surroundings. The system failed to orient the user in the right direction once and one out of the four users asked for assistance once during the experiment. At the end of the stretch, the user was asked to describe his surroundings and the relative positions of the store he walked by. This test aimed at validating our assumption that the users liked exploring and gaining information about their surroundings. We did not intend this test to be a memory game, but most users were able to figure out the relative positions of the stores they walked by with 100% accuracy.

## 6. CONCLUSIONS AND DISCUSSION

We developed an iPhone application, iExplore, which provides assistance in navigating an outdoor shopping complex, for VI users. This application has the potential to turn highly inaccessible locations (like OSC) into blind-friendly and navigable locations. The current version is not yet a perfect solution that solves every issue, but the users from our case study indicated high confidence in exploring the area with the help of this application and relying less on asking for help. We also put forth a summary of some design principles that we learnt through our studies. We then presented a way to obtain the relevant information from multiple sources and use a tiered presentation to effectively present the information. Our study was conducted and the proposed app was tested in one outdoor shopping location. Our framework has the following limitations. Firstly, map registration methods on unseen, non-standardized images provide a relatively low accuracy. But using non-rigid registration helps get an approximate location for each store which can be utilized to give the users a proximate location. Secondly, the store directories available online for download often contain noisy information and legend information that is hard to work with. We worked around this problem by picking our dataset to have moderately complex maps. Our future work will include a detailed study for various layouts and locations. Thirdly, there is a huge element of user subjectivity in our design. We plan to overcome this by increasing the number of subjects in our experiments and testing the application in different OSC locations.

**Acknowledgement**: The work was supported in part by a grant (#1135616) from the National Science Foundation. Any opinions expressed in this material are those of the authors and do not necessarily reflect the views of the NSF.